\pdfoutput=1

\documentclass[11pt]{article}
\usepackage{ragged2e} 
\usepackage{booktabs,makecell, multirow, tabularx}
\usepackage[final]{acl}

\usepackage{times}
\usepackage{latexsym}
\usepackage{pdfpages}

\usepackage[T1]{fontenc}

\usepackage[utf8]{inputenc}

\usepackage{microtype}

\usepackage{inconsolata}

\usepackage{graphicx}
\usepackage{amssymb}
\usepackage{amsmath}
\usepackage{pifont}
\usepackage{booktabs}
\usepackage{array}
\usepackage{subfig}
\usepackage{float}

\title{Beyond Textual Context: Structural Graph Encoding with Adaptive Space Alignment to alleviate the hallucination of LLMs}

\author{
 \textbf{Yifang Zhang\textsuperscript{1}},
 \textbf{Pengfei Duan\textsuperscript{1,*}},
 \textbf{Yiwen Yang\textsuperscript{1}},
 \textbf{Shengwu Xiong\textsuperscript{1,*}}
\\
 \textsuperscript{1}Wuhan University of Technology
\\
 \small{
   \textbf{Correspondence:} \href{mailto:email@domain}{\{duanpf,xiongsw\}@whut.edu.cn}
 }
}

\begin{document}
\maketitle
\begin{abstract}
Currently, the main approach for Large Language Models (LLMs) to tackle the hallucination issue is incorporating Knowledge Graphs(KGs). 
However, LLMs typically treat KGs as plain text, extracting only semantic information and limiting their use of the crucial structural aspects of KGs.
Another challenge is the gap between the embedding spaces of KGs encoders and LLMs text embeddings, which hinders the effective integration of structured knowledge.
To overcome these obstacles, we put forward the \textbf{SSKG - LLM}, an innovative model architecture that is designed to efficiently integrate both the \textbf{\underline{S}}tructural and \textbf{\underline{S}}emantic information of \textbf{\underline{KG}}s into the reasoning processes of \textbf{\underline{LLM}}s. 
\textbf{SSKG - LLM} incorporates the Knowledge Graph Retrieval (KGR) module and the Knowledge Graph Encoding (KGE) module to preserve semantics while utilizing structure. Then, the Knowledge Graph Adaptation (KGA) module is incorporated to enable LLMs to understand KGs embeddings.
We conduct extensive experiments and provide a detailed analysis to explore how incorporating the structural information of KGs can enhance the factual reasoning abilities of LLMs. Our code are available at \url{https://github.com/yfangZhang/SSKG-LLM}.
\end{abstract}

\section{Introduction}

\begin{figure}[!t]
\centering
\includegraphics[width=0.5\textwidth]{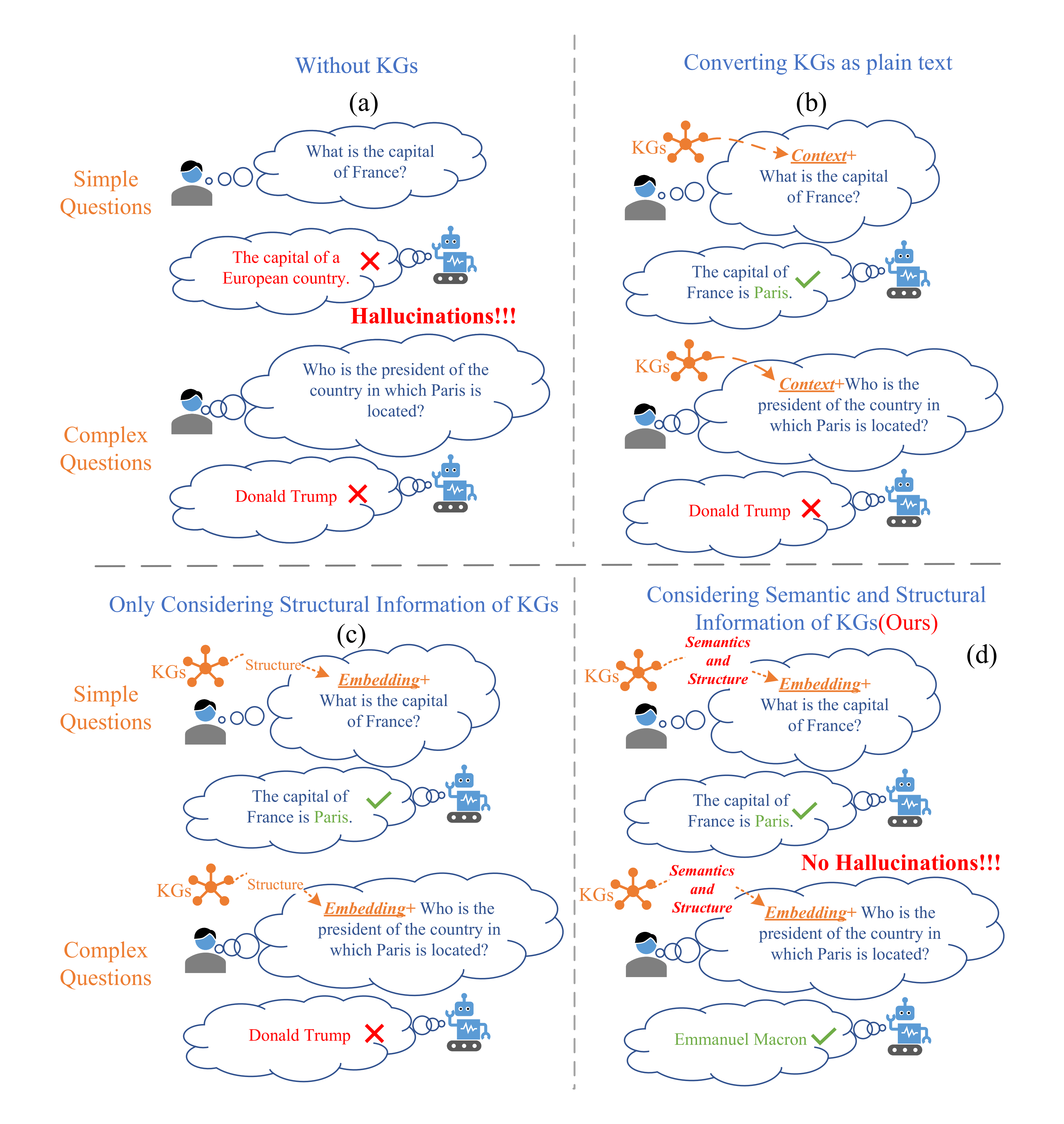}
\caption{The mainstream methods for integrating KGs and LLMs for inference.  
\textit{\textbf{(a) }represents using only LLMs. 
\textbf{(b) }represents converting KGs into plain text.  
\textbf{(c)}represents only incorporating the structural information of KGs.  
\textbf{(d) }represents our proposed method.}} 
\label{fig_intro}
\end{figure}

Recently, LLMs\cite{achiam2023gpt} have demonstrated impressive performance across a range of downstream natural language understanding and generation tasks\cite{du2022glm}. However, despite their remarkable ability to generate fluent and coherent responses, LLMs often struggle with hallucinations and factual inaccuracies, especially when handling knowledge-intensive tasks\cite{pan2024unifying,wan2024acueval}.
As shown in Figure \ref{fig_intro}(a), LLMs can sometimes produce answers that deviate from objective facts.
To address these challenges, researchers propose integrating KGs with LLMs\cite{ijcai2024p734}. KGs store vast amounts of structured knowledge, offering a valuable resource to boost LLMs' knowledge awareness\cite{hu2023survey}.

Mainstream methods for integrating KGs with LLMs fall into three types. The first, exemplified by the GraphRAG series \cite{ijcai2024p734,edge2024local,wen2023mindmap} (see Figure \ref{fig_intro}(b)), serializes the KGs as text and appends it to the user's query. This approach treats the KGs as ordinary text, missing out on its structural benefits.
The second approach uses traditional graph encoders \cite{liu2024explore,luo2024knowla}, like Graph Neural Networks, to encode the KGs, and then combines this with the text encoding before inputting it into the LLMs (see Figure \ref{fig_intro}(c)). However, this method often fails to capture the KGs' full semantic detail and struggles with semantic alignment between the KGs vectors and the LLMs-encoded query, leading to underutilization of the KGs' information. The third method is that LLMs-based KBQA approaches represent an interactive paradigm, where LLMs iteratively explore KGs to build reasoning paths. This approach cannot provide answers to questions not covered by the KGs, limiting the scope of possible answers.

To overcome the shortcomings of the above methods, we propose \textbf{SSKG-LLM}, which enables effective integration of KGs' semantic and structural information during joint reasoning with LLMs. To achieve this, there are two key challenges we need to overcome:

Firstly, 
\textbf{How to acquire and integrate the structural and semantic information of knowledge graphs?} 
KGs store data in a structured, relational format that differs from the text-based representations LLMs are optimized for. They not only include semantic attributes like node labels and entity properties but also capture structural features such as relationships, paths, and hierarchies. Properly incorporating this structured information can significantly enhance the LLMs' understanding of KGs. 

To address this, we first use graph traversal techniques to extract subgraphs relevant to the user's query, ensuring that the intrinsic structure is preserved. Then, we encode the extracted subgraphs with GraphLM\cite{plenz-frank-2024-graph}, a novel pre-trained model that combines the strengths of GNNs and Graph Transformers, thereby retaining both the structural and semantic nuances of the knowledge graph.

Secondly, \textbf{How to bridge the gap between KGs encoding and LLMs?} LLMs are tailored for sequential text processing, while graph encoders produce tensor representations that capture the structured relational data. This fundamental difference in data formats creates a barrier, making it challenging for LLMs to fully leverage the rich information encoded in KGs.

To address this, we propose a KG-Adapter module using cross-attention. It treats graph and text encodings as distinct modalities and aligns them via cross-attention. This integration allows the model to fully utilize KGs information while easing the LLMs' burden in handling multi-modal data, ultimately yielding more accurate answers.

Following the method described above, extensive experiments were conducted on the Multiple types of Question Answering(QA) datasets. Specifically, our approach advances the most models on many QA datasets.

Our main contributions are as follows:

(1) We introduced an innovative framework, SSKG-LLM, for integrating KGs with LLMs. To the best of our knowledge, SSKG-LLM is the first approach to simultaneously integrate both the semantic and structural information of KGs during joint reasoning with LLMs. 

(2) We reveal a misalignment between the representational spaces of KGs encoders and LLMs, which hinders seamless knowledge integration. To bridge this gap, we propose the KGA module that leverages cross-attention.

(3) Experiments on three datasets demonstrate that our method significantly outperforms others in LLMs-based QA tasks with KGs integration, underscoring the importance of structural information.

\section{Related Works}
Integrating LLMs with KGs has advanced QA tasks, with current methods falling into three categories: Hybrid Encoding, GraphRAG, and LLMs-based KBQA.

\textbf{Hybrid Encoding} processes unstructured text and structured KGs separately, then merges them via vector-based integration. For example, KnowLA\cite{luo2024knowla} uses a LoRA-trained layer to embed entity information into LLMs, and the "Explore then Determine"\cite{liu2024explore} framework combines GNNs with LLMs for detailed knowledge extraction. However, these approaches fall short in deeply integrating KGs' semantic and structural details and fail to bridge the gap between graph encoder outputs and LLMs representations.

\textbf{GraphRAG} methods identify the relevant KGs for a question by serializing its triplets into plain text and feeding them along with the query into the LLMs \cite{tang2024graphgpt,ye2023natural,zhao2023graphtext}. For example, MindMap\cite{wen2023mindmap} serializes subgraphs into natural language to boost transparency, while KG-CoT\cite{ijcai2024p734} generates high-confidence reasoning paths to append to the query. Although these methods directly link KGs data to the reasoning process, they often underuse the KGs' inherent relational structure and are sensitive to prompt formatting and context window limitations.

\textbf{LLMs-based KBQA} methods represent an interactive paradigm, with LLMs iteratively navigating KGs to construct reasoning paths\cite{chai2023graphllm,guo2023gpt4graph,
liu2023evaluating,wang2024can}. Methods such as Tree-of-Traversals\cite{markowitz-etal-2024-tree} and Think-on-Graph\cite{sunthink} enable LLMs to act as agents that score and select paths, progressively narrowing down to the optimal solution. While these methods effectively harness the decision-making capabilities of LLMs, they often fail to exploit the models’ direct answering abilities, relying instead on repetitive decision and scoring processes that can introduce inefficiencies. Moreover, their reliance on path exploration highlights a trade-off between stepwise reasoning accuracy and computational cost.

While these methods show progress, they fail to fully leverage the structural and semantic aspects of KGs. Our SSKG-LLM addresses this by efficiently integrating both dimensions into the reasoning process, leading to more accurate and interpretable outcomes.

\section{SSKG-LLM}
Our method addresses two key challenges outlined in the introduction: \textbf{How to acquire and integrate the structural and semantic information of knowledge graphs?} and \textbf{How to bridge the gap between KGs encoding and LLMs?}

To address the first challenge, we designed the \textbf{KGR} and \textbf{KGE} modules. 
(i)The KGR module identifies the subkgs relevant to the user's query and ensures that structural information is preserved while serializing them.
(ii)\textit{KGE}, which encodes the serialized knowledge graph to capture both its semantic and structural information effectively. 

For the second challenge, we specifically tailored the \textbf{KGA} module. (iii)\textit{KGA}, which ensures that the encoded KG is both dimensionally compatible and semantically aligned with LLMs.
\begin{figure}[!t]
\centering
\includegraphics[width=0.5\textwidth]{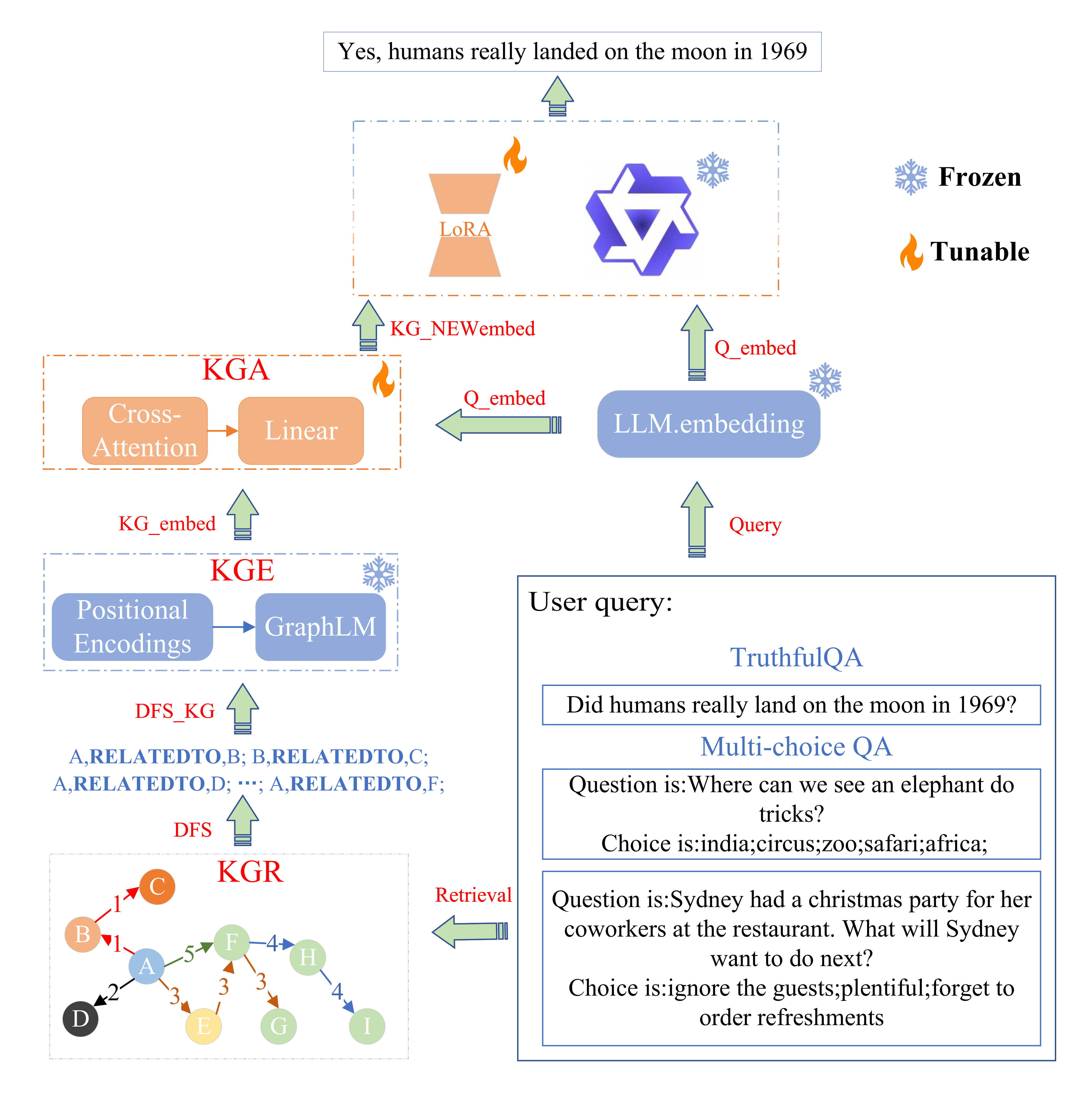}
\caption{Overall Architecture of Models.} 
\label{fig1}
\end{figure}
The overall architecture of the prompt design is detailed in Figure~\ref{fig1}.

\subsection{Knowledge Graph Retrieval}
This module retrieves domain-relevant KGs for a user query through semantic grounding. First, we tokenize the input query $\mathit{query}$ into terms $\{t_1, \dots, t_n\}$ using the SpaCy toolkit, then filter out stopwords via the NLTK stopword list\cite{bird2006nltk}. The remaining keywords are mapped to ConceptNet\cite{speer2017conceptnet} entries through exact string matching, ensuring alignment with commonsense concepts. 
The retrieval process follows three steps:
\begin{equation}
\begin{aligned}
&\{t_1,\cdots, t_n \} = Tokenizer(query)
\end{aligned}
\end{equation}
\begin{equation}
\begin{aligned}
&Topics = GetTopics(\{t_1,\cdots, t_n \})
\end{aligned}
\end{equation}
\begin{equation}
\begin{aligned}
&Sub\_KG = GetKG(Topics)
\end{aligned}
\end{equation}
where $GetTopics$ identifies core concepts via TF-IDF scoring, and $GetKG$ queries Wikidata\cite{vrandevcic2014wikidata} or ConceptNet through their respective APIs.

To enable encoding by graph encoders, the retrieved subgraph is serialized into text. We use a depth-first traversal strategy to preserve the subgraph's hierarchical and relational properties during serialization, ensuring effective downstream encoding and analysis.
\begin{equation}
\begin{aligned}
&s = rand(Topics)
\end{aligned}
\end{equation}
\begin{equation}
\begin{aligned}
&\mathcal{DFS\_KG} = \mathcal{DFS}(Sub\_KG,s)
\end{aligned}
\end{equation}

where $\mathcal{DFS\_KG}$ represents the knowledge graph sequence obtained through depth-first traversal ($\mathcal{DFS}$) starting from node $s$ . $rand(\cdot)$ represents a random selection function. 

\subsection{Knowledge Graph Encoding}
The module encodes KGs by jointly modeling structural and semantic patterns through a frozen language model $GraphLM$. 
\begin{equation}
\begin{aligned}
&KG\_embed = GraphLM(\mathcal{DFS\_KG})
\end{aligned}
\end{equation}

First, we convert the subgraph $\mathcal{DFS\_KG}$ into a Levi graph $Levigraph$ \cite{schmitt-etal-2021-modeling}, where each original edge $(h,r,t)$ becomes a new node $v_r$ connected to $h$ and $t$ via unlabeled edges. This transformation explicitly grounds relational semantics as text-attributable nodes.

Next, a relative position matrix $P$ is constructed based on the resulting Levi graph to encode structural relationships. 

Finally, the serialized Levigraph, along with its corresponding relative position matrix $P$, is input into the T5 model\cite{raffel2020exploring} for pre-training, enabling the model to learn both semantic and structural information from the knowledge graph effectively. And, the embedding representation $KG\_e_{origin}$ that contains both semantic and structural information is obtained. The encoding pipeline involves:
\begin{equation}
\begin{aligned}
&Levigraph = KGs2Levi(\mathcal{DFS\_KG})
\end{aligned}
\end{equation}
\begin{equation}
\begin{aligned}
&P = Positional\_Encoding(Levigraph)
\end{aligned}
\end{equation}
\begin{equation}
\begin{aligned}
&KG\_e_{origin} = T5(Levigraph,P)
\end{aligned}
\end{equation}

This design enables structural reasoning without introducing GNN modules, as empirically validated in \S4

\subsection{Knowledge Graph Adapter}
To bridge the dimensional mismatch and semantic divergence between graph embeddings and LLM input space, we propose a Knowledge Graph Adapter with two stages:

(1) \textbf{Dimensional Alignment}. The linear layer $L$ maps the graph encoder's output vectors to a dimensionality compatible with the LLMs' input space($H\_size$) while preserving their original semantic content. 
\begin{equation}
\begin{aligned}
&KG\_e_{0} = L(KG\_e_{origin},H\_size)
\end{aligned}
\end{equation}

(2) \textbf{Semantic Alignment}. We apllied the cross-attention layer to addresses two critical challenges during semantic alignment: (i) Deeply integrating the encoded knowledge graph representations with the query information. (ii) Bridging the gap between the graph encoder's output and the LLMs by enabling the pre-trained LLMs, which is unfamiliar with graph-encoded vectors, to effectively interpret and utilize this information.
\begin{equation}
\begin{aligned}
&Q\_e = LLMembed(query)
\end{aligned}
\end{equation}
\begin{equation}
\begin{aligned}
& KG\_e = C\_Att(KG\_e_{0}, Q\_e)
\end{aligned}
\end{equation}
where $LLMembed$ denotes the embedding layer of the LLMs, and $C\_Att$ represents the cross-attention layer.

This design ensures seamless interaction between the structured knowledge in the graph and the LLMs' semantic capabilities, enhancing the overall performance on downstream tasks.

\subsection{Fine-tuning LLM}
Subsequently, we employed the Low-Rank Adaptation (LoRA) method to fine-tune the LLMs. LoRA \cite{hu2021lora}integrates low-rank matrices to add trainable adaptation layers without modifying the original model weights, enabling resource-efficient fine-tuning while preserving parameter integrity.

In this section, the KGs embeddings through the KGA layers were concatenated with the text embeddings generated by the LLMs' embedding layer. The resulting concatenated vector was subsequently input into the LoRA-based LLMs fine-tuning module, and the output is Answer $A$. 
\begin{equation}
\begin{aligned}
& I\_e = Concat(KG\_e, Q\_e)
\end{aligned}
\end{equation}
\begin{equation}
\begin{aligned}
&   A = W_{llms}(I\_e) + \Delta W(I\_e)
\end{aligned}
\end{equation}
where, \( Concat \) denotes the vector concatenation function, $I\_e$ is the concatenation of KGs embeddings $KG\_e$ and text embeddings $Q\_e$. \( W_{llms} \) represents the weights of the LLMs, which are kept frozen during the training process. The term \( \Delta W \) refers to the additional parameters introduced during the LoRA fine-tuning process, which are trainable. 

\subsection{Training Objectives}
The auto-regressive training objective focuses on training the LLMs to predict subsequent tokens accurately. Specifically, we calculate the probability of generating the target answer $A$ by:
\begin{equation}
\begin{aligned}
\mathcal{L} &= -\sum_{i=1}^{L} \log P(A_i \mid I\_e, A_{0:i-1}; \theta)
\end{aligned}
\end{equation}
where ${L}$ is the length of the target answer $A$, and
$\theta$ denotes the SSKG-LLM's parameters. The final objective is to minimize the function $\mathcal{L}$. It is important to note that throughout both stages, the weights of both LLMs and the KGE module are unchanged.
\section{Experiments}
\subsection{DataSets}
We consider two types of tasks: multi-choice QA, and truthful QA. We pick CommonsenseQA\cite{talmor2019commonsenseqa} and SIQA\cite{sap2019social} as the multiple-choice QA datasets. We also use TruthfulQA\cite{lin2022truthfulqa} to evaluate whether SSKG-LLM is truthful in generating answers to questions.

For the two multiple-choice QA datasets, we use accuracy (ACC) as the evaluation metric. A prediction is considered correct only if the model's output matches the ground truth exactly.
For the TruthfulQA dataset, we adopt commonly used similarity metrics, such as ROUGE-N\cite{lin2004rouge} and BLEU\cite{papineni2002bleu}, and use their average score as the final experimental result to ensure a comprehensive evaluation of the model's performance.
\subsection{Experiments Settings}
We consider the following LLMs with 13B or 14B parameters as foundation models in our main experiments:

we fine-tune the Qwen1.5-14B-Chat\cite{qwen} \footnote{\url{https://huggingface.co/Qwen}}  proposed by Alibaba Cloud and Llama2-13B-Chat\cite{touvron2023llama} developed by Meta which has 13B parameters. 

A machine equipped with a NVIDIA GeForce RTX 5880 GPU (with 48GB of VRAM) is used for training. The learning rate is 1e-4 with linear warm-up for the first 100 training steps and the batch size is set to 4. During the training process, we use the AdamW optimizer.

\subsection{Results}
To demonstrate the effectiveness of our proposed method, we compare it with several existing models for LLMs-based question-answering tasks with knowledge graph integration. 

To highlight the comprehensiveness of the comparison, we contrast our proposed method with the existing four major methods of integrating KGs and LLMs for inference. The existing four major methodss are as follows: (1) Methods that \textbf{only LLM} for inference. (2) Methods that convert \textbf{knowledge graphs into plain text} to serve as prompts. (3) \textbf{LLMs-based KBQA}, which involve LLMs iteratively navigating KGs to construct reasoning paths. (4) \textbf{Hybrid encoding} methods.

• \textbf{BaseLLM(No-retrieval)} evaluates the performance of base LLMs, where the inputs do not contain any retrieval content of KGs.

• \textbf{BaseLLM(RAG)}  evaluates the performance of base LLMs, where the inputs contain the
retrieval KGs.

• \textbf{LLM-SFT(No-retrieval)} evaluates the performance of continual pre-trained
LLMs after LoRA-based SFT with the constructed
training set, where the inputs do not contain any retrieval content of KGs.

• \textbf{LLM-SFT(RAG)} evaluates the performance of continual pre-trained
LLMs after LoRA-based SFT with the constructed
training set, where the inputs do contain the
retrieval KGs.

• \textbf{GPT-3.5 + KSL}\cite{2023Knowledge} designs a effective prompt
to transform retrieval KGs into a multi-hop decision sequence, which empowers LLMs with searching
knowledge ability.

• \textbf{GPT-4\&ChatGPT + ToG-R}\cite{sun2024thinkongraph} instructs LLM itself to perform retrieval, pruning and answer prediction on KGs.

• \textbf{KG-COT(ChatGPT)}\cite{ijcai2024p734} leverages a small-scale step-by-step graph reasoning model to reason over KGs and utilizes a reasoning path generation method to generate chains of knowledge with high confidence for LLMs

• \textbf{KnowLA(ConceptNet)}\cite{luo2024knowla} proposes a knowledgeable adaptation method. It inserts an adaptation layer into an LLM to integrate the embeddings of  entities of KG appearing in the input text.

• \textbf{KAPING (ConceptNet)}\cite{baek-etal-2023-knowledge} retrieves rele-
vant triples from KGs to improve the KBQA
task. We use KAPING to enhance LLMs on
knowledge-relevant tasks.

• \textbf{KG-Adapter}\cite{tian2024kg} proposed a parameter-level
KG integration method based on parameter efficient fine-tuning.
\begin{table*}[!t]
    \centering
    \caption{Results of SSKG-LLM (\textbf{Bold} values indicate the best model, while \underline{underlined} represents the second-best.)}
   {
    \resizebox{\linewidth}{!}{ 
    \begin{tabular}{ccccccc}
    \hline 
     Method & 
     Model&
    CommonsenseQA & 
    SIQA & 
    \multicolumn{3}{c}{TruthfulQA }\\
    \cline{5-7}
    &&$\mathrm{ACC}$ &$\mathrm{ACC}$ &$\mathrm{ROUGE-1}$ & $\mathrm{ROUGE-2}$ & $\mathrm{BLEU-score}$ \\
    \hline \multicolumn{1}{c}{\multirow{4}{*}{Only LLM}}&Qwen1.5-14B(No-retrieval)&78.37&71.79&0.1944&0.0973&0.0438\\
           &Llama2-13B(No-retrieval)&78.56&71.85&0.1985&0.0997&0.0488\\
           &Qwen1.5-14B-SFT(No-retrieval)&85.83&76.46&0.5078&0.3752&0.2903\\
           &Llama2-13B-SFT(No-retrieval)&85.65&76.36&0.5098&0.3762&0.2923\\
    \hline
            \multicolumn{1}{c}{\multirow{5}{*}{KGs to Plain-text}}&Qwen1.5-14B(RAG)&70.76&71.02&0.2232&0.1168&0.0529\\
           &Llama2-13B(RAG)&70.81&71.22&0.2223&0.1157&0.0515\\
           &Qwen1.5-14B-SFT(RAG)&85.59&77.18&0.5518&0.4252&0.3391\\
           &Llama2-13B-SFT(RAG)&85.56&77.13&0.5516&0.4235&0.3371\\
           &GPT-3.5 + KSL&79.60&-&-&-&-\\
           
    \hline \multicolumn{1}{c}{\multirow{3}{*}{LLMs-based KBQA}}&ChatGPT + ToG-R&-&77.8&-&-&-\\
    &GPT-4 + ToG-R&-&45.4&-&-&-\\
    &KG-CoT(ChatGPT)&-&58.6&-&-&-\\
    \hline
           \multicolumn{1}{c}{\multirow{3}{*}{Hybrid Encoding(only semantics)}}&KnowLA (ConceptNet)&58.39&53.22&0.4190&0.2922&0.1747\\
           &KAPING (ConceptNet)&57.58&52.66&-&-&-\\
           &KnowLA (WordNet)& 58.07&53.22&0.4143 &0.2874 &0.1714\\
    \hline
           Hybrid Encoding(only structures)&KG-Adapter&79.60&-&-&-&-\\
    \hline
           \multicolumn{1}{c}{\multirow{2}{*}{\bfseries Ours}}&\bfseries SSKG-LLM(Ours-Llama2-13B)& \bfseries86.54& \underline{77.92}& \underline{0.5582}& \underline{0.4315}& \underline{ 0.3459}\\
    
 & \bfseries SSKG-LLM(Ours-Qwen1.5-14B)& \underline{86.32} & \bfseries77.95& \bfseries0.5585 & \bfseries0.4317&\bfseries 0.3462
\\
\hline
    \end{tabular}
    }
    }
    \label{tab_multi-choice}
\end{table*}

As shown in Table \ref{tab_multi-choice}, SSKG-LLM consistently outperforms all baseline methods across both multiple-choice QA datasets and the TruthfulQA dataset. The improvements across different datasets are approximately 8\%, demonstrating the effectiveness of our proposed method. This indicates that SSKG-LLM not only leverages the semantic and structural information from knowledge graphs to generate more accurate answers for multiple-choice QA tasks but also produces responses that align better with common knowledge and are more trustworthy.

In the multiple-choice QA task, treating knowledge graphs as plain text and using the RAG method, regardless of fine-tuning, results in lower performance compared to the no-retrieval setting. This is because multiple-choice tasks prioritize precision, and excessive textual information leads to redundancy, which hampers decision-making. In contrast, SSKG-LLM leverages both the semantic and structural aspects of knowledge graphs for high-dimensional vector representation, enhancing accuracy without introducing redundancy.

As shown in Table \ref{tab_multi-choice}, SSKG-LLM significantly outperforms the no-retrieval setting on the TruthfulQA dataset for open-ended questions. Under the RAG setting, the similarity between SSKG-LLM's generated answers and the true answers is higher, improving credibility, as the TruthfulQA dataset evaluates answer truthfulness. This is due to the knowledge graph providing valuable reference information, allowing the LLMs to focus on generating trustworthy answers rather than achieving complete accuracy. Moreover, by incorporating the knowledge graph's structural information into the reasoning process, SSKG-LLM significantly outperforms the RAG method alone.
\subsection{Results Analysis}
\subsubsection{What Are the Roles of Each Module in the SSKG-LLM?}
\begin{table}[!h]
    \centering
    \caption{Results of Ablation}
    \resizebox{\linewidth}{!}{
    \begin{tabular}{cccccc}
    \hline 
    \multicolumn{3}{c}{Module} & \multirow{2}{*}{CommonsenseQA} & \multirow{2}{*}{SIQA} & \multirow{2}{*}{TruthfulQA} \\
    \cline{1-3} 
    $\mathrm{KGR}$ & $\mathrm{KGE}$ & $\mathrm{KGA}$ &  &  &  \\
    \hline
    \checkmark & \checkmark & \ding{55} & 85.34 & 70.77 & 0.3388 \\
    \checkmark & \ding{55} & \ding{55} & 85.59 & 77.18 & 0.3391 \\
    \ding{55} & \ding{55} & \ding{55} & 78.37 & 71.79 & 0.0438 \\
    \checkmark & \checkmark & \checkmark & \bfseries86.32 & \bfseries77.95 & \bfseries0.3462 \\
    \hline
    \end{tabular}
    }
    \label{tab_ablation}
\end{table}

To validate the effectiveness of the proposed method in this paper, we conducted corresponding ablation experiments to assess the impact of each module on the SSKG-LLM's performance.

• In the KGR column, the \checkmark indicates the use of the method described in Section 3.1, while the \ding{55} denotes that no knowledge graph extraction is performed, meaning the input to the LLMs consists solely of the question.

• In the KGE column, the \checkmark indicates the use of the method described in Section 3.2, while a \ding{55} signifies that no knowledge graph encoding is performed. Instead, the KGs are treated as plain text and directly input into the LLMs along with the question, without additional processing.

• In the KGA column, the \checkmark indicates the use of the model architecture described in Section 3.3, while the \ding{55} denotes that only the linear layer from Section 3.3 is retained, with all other components removed. The KG embeddings are directly concatenated with the query embeddings, which is used as input to the LLMs.

As shown in Table \ref{tab_ablation}, removing the KGA module leads to a performance decline, particularly on the SIQA dataset, where accuracy drops by 7.18\%. Furthermore, when the KGE module is removed, the model's performance also deteriorates. 
Finally, when all three modules are removed, there is a significant decline in the model's performance.
\subsubsection{Does the Gap Exist Between KGs and LLMs' Embedding Spaces, and How Can It Be Bridged?}
Interestingly, we observe that the removal of both the KGA and KGE modules leads to a slight improvement, compared to the removal of only the KGA module (1st and 2nd rows of Table \ref{tab_ablation}).

Through analyzing the experimental data, we attribute this phenomenon to the following: when the KG embedding generated by the KG encoder is directly input to the LLMs without adaptation, it will exacerbate the gap between the LLMs and the KG encoder, as discussed in the Introduction. However, when the knowledge graph is treated as plain text rather than using the graph encoder, this gap naturally disappears.

This observation provides evidence in two aspects. Firstly, it confirms the existence of a gap between LLMs and KG encoders. Secondly, it demonstrates that the proposed KG-Adapter not only mitigates this gap to a significant extent but also leverages the rich semantic and structural information contained in KGs effectively.

\subsubsection{Why structural information of KGs is important and must not be overlooked?}
To investigate the significance of structural information in knowledge graphs and demonstrate the advantages of our proposed method in effectively integrating both semantic and structural information for enhanced utilization by LLMs, we designed the following two experiments:

\textbf{(1)What Are the Effects of Graph Traversal Strategies?}
We systematically evaluated different graph traversal methodologies within our KGR framework to assess their effectiveness in preserving structural information from knowledge graphs.



• \textbf{DFS (Depth-First Search)} \cite{kozen1992depth} traverses graphs by following a single branch as deep as possible before backtracking, effectively uncovering long chains and generating hierarchical, logical knowledge graph sequences.

• \textbf{BFS (Breadth-First Search)} \cite{kozen1992depth} explores graphs layer by layer, offering broad coverage and capturing information near the starting node, though it tends to produce shorter chains and may miss longer-range dependencies.

• \textbf{Random Walk} selects neighboring nodes at random at each step, enabling diverse path exploration but potentially hindering the systematic discovery of critical paths, which can result in less structured and coherent knowledge graph outputs.

\begin{figure}[!h]
\centering
\includegraphics[width=0.5\textwidth]{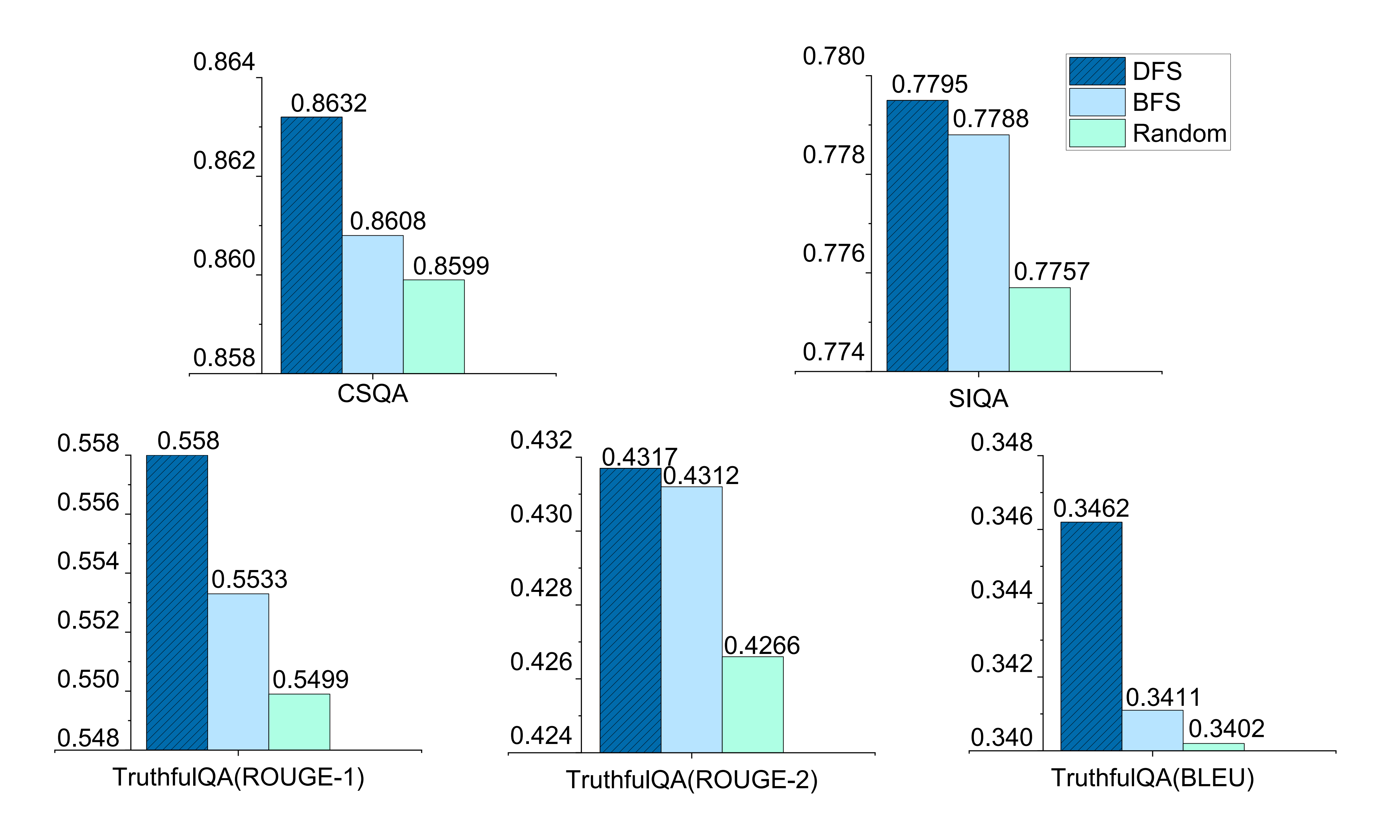}
\caption{Results of Varying Graph Traversal Strategies.}
\label{fig_Traversal_truthfulqa}
\end{figure}

As shown in Figure \ref{fig_Traversal_truthfulqa}, the DFS algorithm demonstrated the best performance, while the Random Walk method showed the poorest results. 

Upon analyzing the content of the generated knowledge graph sequences and the corresponding experimental outcomes, we found that the superior performance of DFS can be attributed to its ability to generate longer knowledge graph chains. These chains are similar to Chain-of-Thought (COT) reasoning chains, a format that has been shown to enhance the reasoning abilities of  LLMs in various domains\cite{feng2024towards,ranaldi2024aligning}, leading to more accurate answers.

\textbf{(2)What Are the Effects of Using Various Knowledge Graph Encoders?}
Subsequently, we conducted comprehensive experiments to examine the performance of different KGs embedding approaches in capturing and integrating both semantic and structural information. This investigation focuses on measuring the embedding techniques' capacity to preserve intrinsic relationships while effectively fusing heterogeneous information types.
To compare the impact of different knowledge graph encoders on the results, we selected several commonly used graph encoders, including GNN, GraphTransformer, and the GraphLM used in SSKG-LLM. The experiment was conducted with the following four settings:

\begin{figure}[!h]
\centering
\includegraphics[width=0.3\textwidth]{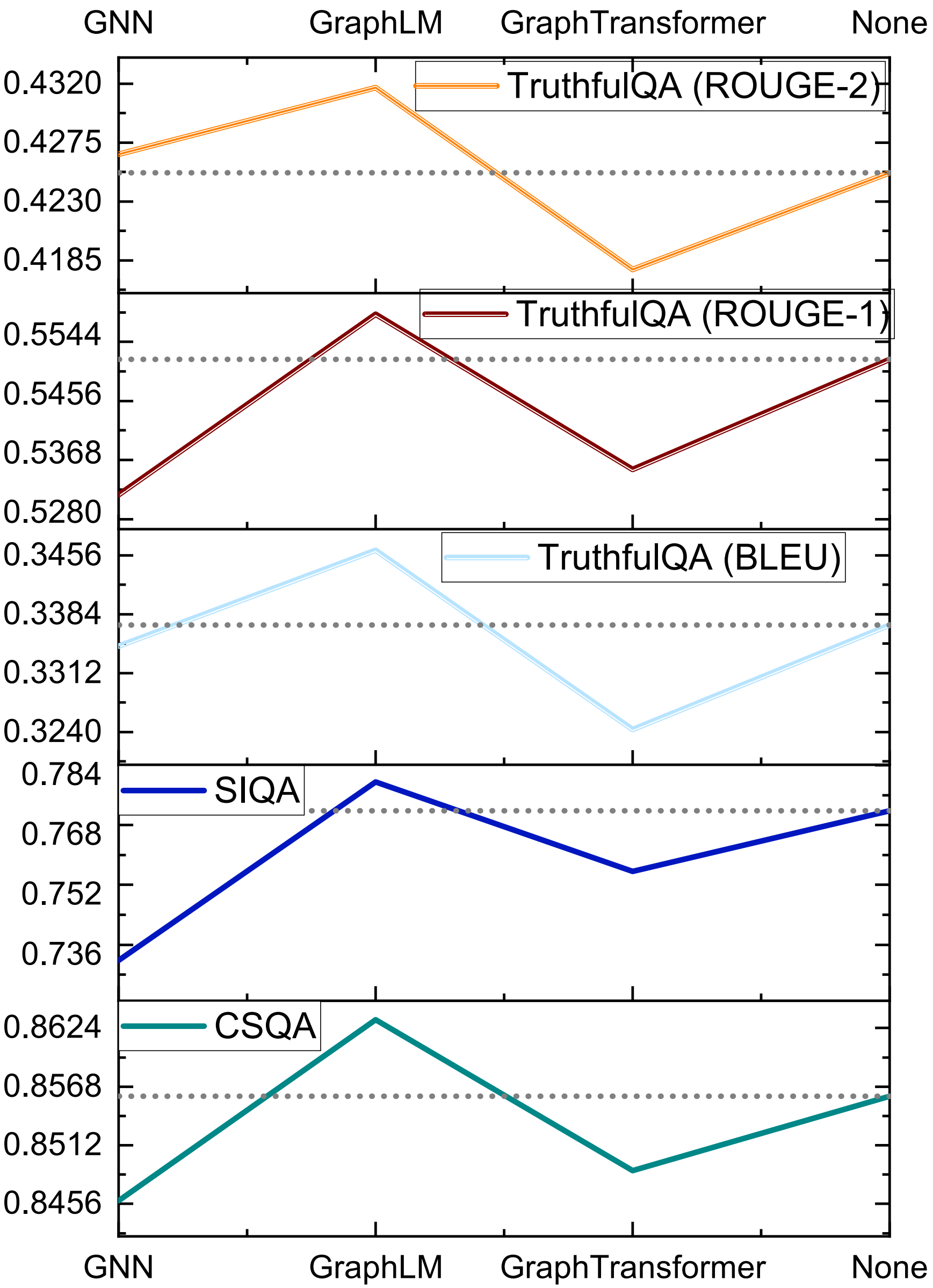}
\caption{Results of using
various Knowledge Graph Encoders.}
\label{fig_graphencoder}
\end{figure}

• \textbf{GNN}\cite{zhou2020graph} is a class of neural networks specifically designed to work with graph-structured data. It operates by aggregating information from neighboring nodes to update the representation of each node, allowing it to capture local dependencies and structural information in the graph.

• \textbf{GraphTransformer}\cite{yang2021graphformers} is an extension of the Transformer model designed for graph data. It utilizes self-attention mechanisms to model long-range dependencies in the graph.

• \textbf{GraphLM}\cite{plenz-frank-2024-graph}, the encoder used in SSKG-LLM, is specifically designed to encode knowledge graphs in a way that efficiently captures both structural and semantic information. It combines the advantages of both GNN and Transformer architectures, making it particularly effective for tasks that require rich, hierarchical graph representations.

• \textbf{None}: This setting refers to the baseline case where no knowledge graph encoding is applied, and the model only uses the query for inference.

As illustrated in Figure \ref{fig_graphencoder}, GraphLM consistently delivers superior performance, demonstrating its ability to capture both structural and semantic information in KGs, which is essential for tasks requiring nuanced reasoning and accurate generation.

Unlike GNN, which focuses primarily on local structure by aggregating information from neighboring nodes, GraphLM integrates these structural features with a broader semantic understanding. This capability allows it to overcome GNN’s limitations in handling long-range dependencies, ensuring better performance in tasks that rely on deeper graph comprehension. At the same time, GraphLM addresses the shortcomings of GraphTransformer, which excels in modeling semantic relationships but struggles to represent the hierarchical structure of graphs effectively. By balancing these two aspects, GraphLM achieves a more comprehensive encoding of KGs, making it highly effective across diverse tasks.


GraphLM's consistent outperformance highlights the value of hybrid architectures that fuse structural and semantic encoding. By jointly leveraging local and global graph information, SSKG-LLM fully exploits the potential of knowledge graphs, surpassing alternatives such as GNNs and Graph Transformers. This demonstrates that both structural and semantic attributes are essential for optimal utilization of knowledge graphs.

\section{Conclusion}
In this paper, we introduce SSKG-LLM, a novel architecture that efficiently integrates both the structural and semantic information of KGs into LLMs reasoning. Extensive experiments validate its effectiveness across diverse tasks, highlighting the importance of leveraging both KGs' aspects for more accurate and robust reasoning.

\section*{Limitation}
While our method enables LLMs to achieve strong performance across three question-answering datasets, several limitations remain. First, despite being parameter-efficient, our approach still demands more computational resources than methods relying solely on LLMs inference. Additionally, our experiments are limited to English datasets, restricting our ability to determine whether these issues persist in other languages. We will address these limitations in our future work.

\bibliography{custom}

\appendix

\section{Robustness of SSKG-LLM}
\label{sec:appendix}
To assess the robustness of the proposed SSKG-LLM, a comprehensive series of experiments was conducted, evaluating its performance across multiple datasets and varying configurations of Low-Rank Adaptation (LoRA) parameters. This analysis was crucial to demonstrate the stability and generalizability of SSKG-LLM when subjected to diverse fine-tuning conditions. Specifically, we examined the model's performance using three distinct rank values: 
R=8, R=16, and R=32, during the LoRA fine-tuning process. These values represent different levels of parameter efficiency, where lower ranks correspond to more compact and computationally efficient parameterizations, while higher ranks provide greater capacity for capturing nuanced patterns in the data.
\begin{figure}[!h]
\centering
\includegraphics[width=0.5\textwidth]{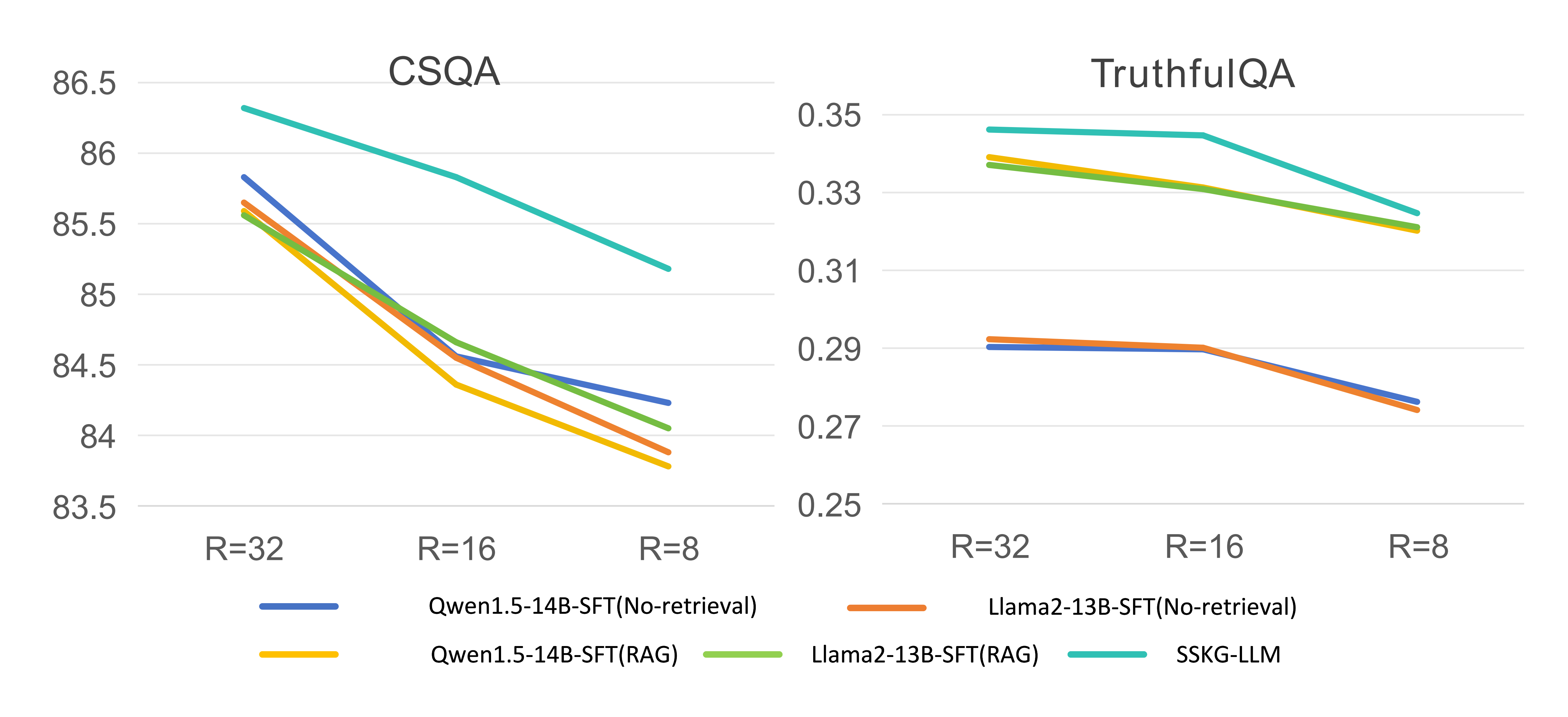}
\caption{Results of Varying LoRA parameter settings.}
\label{fig_lora}
\end{figure}

The experimental results, depicted in Figure \ref{fig_lora}, span two datasets CSQA and TruthfulQA and include comparisons against several baseline configurations, including Qwen-1.5-14B-SFT and Llama2-13B-SFT. The baselines were chosen to ensure a comprehensive evaluation across different model architectures, fine-tuning strategies, and dataset types.

\subsubsection{Model-Level Analysis}
The proposed framework proves highly adaptable across different underlying LLMs architectures. When applied to Qwen-1.5-14B and Llama2-13B, SSKG-LLM consistently outperforms baseline settings for both models, including those enhanced with retrieval-augmented generation (RAG). This adaptability underscores the versatility of the SSKG-LLM framework, as it can be seamlessly integrated with different pretrained LLMs while maintaining robust and superior performance. Unlike baselines, which exhibit variability based on the specific model and task, SSKG-LLM delivers consistently strong results, demonstrating its universal applicability across architectures.

\subsubsection{Dataset-Level Analysis} The experimental results demonstrate that SSKG-LLM exhibits superior performance across both the CSQA and TruthfulQA datasets, highlighting its versatility in addressing diverse task demands. On CSQA, which emphasizes commonsense reasoning, SSKG-LLM consistently outperforms baseline models under all LoRA parameter settings. Moreover, its performance degrades more gracefully as the rank decreases from R=32 to R=8, underscoring its ability to effectively leverage structured knowledge for complex reasoning tasks. Similarly, on TruthfulQA, a dataset focused on factuality and truthfulness, SSKG-LLM achieves both higher accuracy and greater robustness across all ranks, even at the most constrained configuration (R=8). This cross-dataset robustness suggests that SSKG-LLM is not only task-agnostic but also capable of maintaining high-quality performance in both reasoning-intensive and knowledge-intensive tasks, making it a highly adaptable solution for diverse NLP challenges.

\subsubsection{Parameter-Level Analysis}
SSKG-LLM exhibits remarkable robustness under different LoRA ranks, with significantly less performance degradation as rank decreases from R=32 to R=8. At R=8, where baselines suffer notable drops, SSKG-LLM sustains competitive results, demonstrating its efficiency in low-resource scenarios. This resilience highlights SSKG-LLM’s practicality for applications requiring computational efficiency while maintaining high-quality performance.

As shown in Figure \ref{fig_lora}, SSKG-LLM consistently outperforms the baseline methods across all parameter settings. This experiment demonstrates that SSKG-LLM is robust and can lead to stable improvements when combined with different LLMs, datasets, and ranks.

\end{document}